\documentclass[conference]{IEEEtran}
\ifCLASSINFOpdf
\else
\fi

\usepackage[numbers]{natbib}
\usepackage{graphicx}
\usepackage{comment}
\usepackage{hyperref}
\usepackage{multirow}
\usepackage{multicol}
\usepackage{amsmath}

\begin{document}
	%
	\title{HSD Shared Task in VLSP Campaign 2019:\\
		Hate Speech Detection for Social Good}

	
	
	%
	\author{\IEEEauthorblockN{Xuan-Son Vu\IEEEauthorrefmark{1},
			Thanh Vu\IEEEauthorrefmark{2},
			Mai-Vu Tran\IEEEauthorrefmark{3}, 
			Thanh Le-Cong\IEEEauthorrefmark{4}, 
			Huyen T M. Nguyen\IEEEauthorrefmark{5} 
		}
		\IEEEauthorblockA{\IEEEauthorrefmark{1} Deep Data Mining Group, Department of Computing Science, Ume\r{a} University,
			Sweden. Email: sonvx@cs.umu.se}
		\IEEEauthorblockA{\IEEEauthorrefmark{2}The Australian E-Health Research Centre, CSIRO, Australia. Email: thanh.vu@csiro.au}
		\IEEEauthorblockA{\IEEEauthorrefmark{3}University of Engineering and Technology (UET), 
			Vietnam National University, Hanoi. Email: vutm@vnu.edu.vn}
		\IEEEauthorblockA{\IEEEauthorrefmark{4}InfoRE Technology Company. Email: thanh@infore.vn}
		\IEEEauthorblockA{\IEEEauthorrefmark{5}Hanoi University of Science (HUS), Vietnam National University, Hanoi. Email: huyenntm@hus.edu.vn}}


	\maketitle
	
	\begin{abstract}
		
		
		
		The paper describes the organisation of the ``Hate Speech Detection'' (HSD) task at the VLSP workshop 2019 on detecting the fine-grained presence of hate speech in Vietnamese textual items (i.e., messages) extracted from Facebook, which is the most popular social network site (SNS) in Vietnam.  The task is organised as a multi-class classification task and based on a large-scale dataset containing 25,431 Vietnamese textual items from  Facebook. The task participants were challenged to build a classification model that is capable of classifying an item to one of 3 classes, i.e., ``HATE', ``OFFENSIVE'' and ``CLEAN''. HSD  attracted a large number of participants and was a popular task at VLSP 2019. In particular, there were 71 teams signed up for the task, 14 of them submitted results with 380 valid submissions from 20$^{th}$ September 2019 to 4$^{th}$ October 2019.
	\end{abstract}
	

	%
	\IEEEpeerreviewmaketitle

	\section{Introduction}
	
	
	
	On social network sites (SNSs), such as Facebook, Twitter, the threat of abuse and harassment online makes many SNS users stop expressing themselves as well as seeking different opinions. This problem is not trivial to be handled. And SNSs have been struggling with it. For example, to overcome the problem, SNSs might limit or even completely shut down the user post/comment functions in some communities (groups) producing ``not-clean'' content. This, however, further creates an issue of blocking ``clean'' content produced by the same communities. 
	
	To handle the problem, one of the popular strategies is to train systems capable of recognising hateful (``not-lean'') contents, which can then be removed or quarantined by the moderators of communities. In the last few years, much attention has been paid to the problem of detecting hateful contents in SNSs \cite{davidson2017automated, wiegand2018overview, kumar-etal-2018-benchmarking, basile2019semeval}. However, the research is focused mainly on popular languages, such as English \cite{xu2012learning, davidson2017automated, kumar-etal-2018-benchmarking, basile2019semeval}. Despite the large number of SNS users in Vietnam expected to reach 48 million users by the end of 2019\footnote{\url{http://bit.ly/number-of-social-network-users-in-vietnam/}}, to our knowledge, there is no publicly available research on hate speech detection for Vietnamese. 
	
	To this end, we first introduce the task of hate speech detection (HSD) in SNSs for Vietnamese with the aim of supporting more effective conversations in SNSs. The task is organised at the VLSP 2019, which is the sixth annual international workshop in conjunction with the 2019 Conference of the Pacific Association for Computational Linguistics (PACLING 2019).
	
	
	The remainder of the paper is as follows. In the next section,  the data collection and annotation methodologies are described. The shared task description and evaluation are summarised in Section \ref{eval}. Section \ref{part} describes the participants and results. Section \ref{con} concludes the paper as well as shows possible designs for the next year challenge.
	
	\section{Data Collection and Annotation}
	\label{dataset}
	A general corpus is firstly collected from Facebook posts and comments. 
	From the general corpus, we built a neural model to select about 25,431 items for manual annotation. We proposed a pipeline to select these 25,431 items, we as follows:
	\begin{itemize}
		\item Based on top obscene keywords\footnote{\url{https://github.com/vietnlp/vlsp2019_hatespeech_task/}} in Vietnamese, we apply semantic search to find 200 most relevant items in the collected corpus.
		\item Annotators were asked to initially annotate the above 200 items (i.e., to label each of them to one of three classes: hate speech (HATE), offensive but not hate speech (OFFENSIVE), neither offensive nor hate speech (CLEAN).
		\item Based on the above 200 annotated items, we built a classifier to predict the chance of every item in the BIG corpus belongs to each of 3 classes.
		\item Top $K$ items for each class is selected until we reach a total of 25,431 items. Normally, many items will belong to the \emph{CLEAN} class, therefore, we prioritise items belong to less popular classes (i.e., \emph{\{HATE,  OFFENSIVE\}}) unless if any of them has more than 9,000 items (i.e., more than $\frac{1}{3}$ of $25,431$).
		
	\end{itemize}
	
	\subsection{Data Annotation}
	
	From the above initial 25,431 items, we ask twenty-five annotators to manually annotate them in one month. Each item was annotated by three annotators to label each item as one of three categories: hate speech (HATE), offensive but not hate speech (OFFENSIVE), or neither offensive nor hate speech (CLEAN). The annotators were provided with our pre-defined annotation guideline, in which each category is associated with a definition and a paragraph explaining the definition in detail. The annotators were asked to consider not only terms (words) appearing in a given item but also about the context in which they thought the terms (words) were used. The annotators were also instructed that the presence of particular words, such as offensive words, does not necessarily indicate the corresponding item is hate speech. Since each item is annotated by three annotators, we used the majority voting schema to decide the final label of the item.
	
	Here is the detail explanation for each type of three classes:
	
	\begin{itemize}
		\item \textbf{Hate speech (HATE)}: an item is identified as hate speech if it (1) targets individual or groups on the basis of their characteristics; (2) demonstrates a clear intention to incite harm, or to promote hatred; (3) may or may not use offensive or profane words. For example: ``Assimilate? No they all need to go back to their own countries. \#BanMuslims Sorry if someone disagrees too bad.". See the definition of (see definition of Zhang et al.~\cite{Zhang:2018}). In contrast, ``All you perverts (other than me) who posted today, needs to leave the O Board'' is an example of abusive language, which often bears the purpose of insulting individuals or groups, and can include hate speech, derogatory and offensive language.
		\item \textbf{Offensive but not hate speech (OFFENSIVE)}: an item (posts/comments) may contain offensive words but it does not target individual or groups on the basis of their characteristics. E.g., ``WTF, tomorrow is Monday already?''
		\item \textbf{Neither offensive nor hate speech (CLEAN)}: normal item, it does not contain offensive languages or hate speech. E.g., ``She learned how to paint very hard when she was young''.
	\end{itemize}
	
	\subsection{Data Pre-processing}
	As the data might contain sensitive information such as email address, phone number, we run data pre-processing to remove or anonymise the sensitive information. Here is the list of pre-processed information in the user posts/comments:
	
	\begin{enumerate}
		\item All links are replaced by \emph{$<$URL$>$}.
		\item Three last digits of phone numbers are replaced by \emph{XXX}.
		\item The first part of email addresses are replaced by \emph{AAA}.
	\end{enumerate}
	
	Although we tried to anonymise sensitive information. The data itself is very sensitive. Therefore, we stated that by joining the challenge, all participants are not allowed to attempt to \emph{re-}identify the owner of any post or comment in any form or circumstance.

	\section{Shared Task Description and Evaluation}
	\label{eval}
	In this shared task, participants are challenged to build a multi-class classification model that is capable of classifying an item to one of three classes (HATE, OFFENSIVE, CLEAN). The prepared dataset was provided to all participants. The data were randomly split into two parts: the training data and the test data. The test data contains both ``public-test'' and ``private-test''. The public-test was used to allow all participated teams to tune their proposed models. They could submit at most five submissions per day. The final ranking was based on the private-test set. The private-test set was used to ensure the predictive models were not over-fit on the training data and hence, perform equally well on the private-test data. The evaluation metric used in the shared task is the macro-averaged F1 score (Macro-F1). The metric is calculated as follows:
	\begin{equation*}
		\textit{Macro-F1} = \frac{\textit{F1(HATE) + F1(OFFENSIVE) + F1(CLEAN)}}{3}
	\end{equation*}
	
	\noindent Here,
	\begin{equation*}
		\textit{F1} = \frac{\textit{2 x Precision x Recall}}{\textit{Precision + Recall}}
	\end{equation*}
	
	\begin{equation*}
		\textit{Precision} = \frac{\textit{number of correctly predicted instances}}{\textit{number of predicted labels}}
	\end{equation*}
	
	\begin{equation*}
		\textit{Recall} = \frac{\textit{number of correctly predicted labels}}{\textit{number labels in the gold standard}}
	\end{equation*}

	\section{Participants and Results}
	\label{part}
	There are 71 teams registered for this year's challenge and 35 ones that obtained the data after sending the signed user agreement. Finally, only 14 teams participated with 380 submissions during the period of 14 days from 20$^{th}$ September 2019 to 04$^{th}$ October 2019. The performances of the top five teams on the public-test and the private-test are detailed in Table~\ref{table1:top5participants}. It can be seen that the average performance of the top-5 teams on the public-test of the top five participated teams is about +12.5\% absolute higher than that on the private-test. Moreover, although the public-test and private-test are distributed differently. There are 5 out of 8 teams stay in the top-8 of both public-test and private test. This means that competing to achieve higher scores on the public-test with the expectation of getting higher performances on the private-test is still hold even with the highly different distributions of the public-test and private-test data. 
	
	\begin{table}[]
		\caption{Top 5 teams on public-test and private-test. Evaluation metric is Macro-F1.}
		\centering
		\label{table1:top5participants}
		\begin{tabular}{|l|ll|ll|l}
			\hline
			\# & Public-Test & Macro-F1 & Private-Test & Macro-F1      \\ \hline
			1  & Try hard & 0.73019  & SunBear (1st place) & 0.61971 \\ \hline
			2  & HH\_UIT  & 0.71432  & ABCD (2nd place)    & 0.58883 \\ \hline
			3  & titanic  & 0.70747  & Try hard (3rd place) & 0.58455 \\ \hline
			4  & ABCD     & 0.70582  & Cr4zy (on-hold) &    0.57357 \\ \hline
			5  & TIN HUYNH & 0.70576 & BA (on-hold)   &  0.56281    \\ \hline
			
			- & \emph{Top-5 Average} & 0.71271 & \emph{Top-5 Average} & 0.58589 \\
			\hline
		\end{tabular}
	\end{table}
	
	\begin{table*}[]
		\caption{Top 5 teams on public-test and private-test with submitted papers and their final approaches. The rank is based on the macro-averaged F1 scores on the private-test.}
		\centering
		\label{table2:approaches}
		\begin{tabular}{|l|p{2.5cm}|l|l|p{6.0cm}|l|l|}
			\hline
			\multirow{2}{*}{\#} & \multirow{2}{*}{Team} & \multicolumn{2}{c|}{Macro-F1} & \multirow{2}{*}{Final Approach}  & \multirow{2}{*}{Ensemble?} & \multirow{2}{*}{Deep learning?}       \\ \cline{3-4}
			& & Public-test & Private-test &   & &      \\ \hline
			1 & SunBear (1$^{st}$ place) & 0.67756 & \emph{0.61971} & Logistic Regression (LR) & Yes & No\\ \hline
			2 & ABCD (2$^{nd}$ place) & 0.70582 & 0.58883 &  LR, Extra Trees, Random Forest. & Yes & No \\ \hline
			3 & Try hard (3$^{rd}$ place) & \emph{0.73019} & 0.58455 & VDCNN, TextCNN, LSTM, LSTMCNN, SARNN & Yes & Yes \\ \hline
			4 & HH\_UIT  & 0.71432 & 0.56281 &    Bi-LSTM  & No & Yes\\ \hline
			5 & TIN HUYNH & 0.70576    & 0.51705  & Bi-GRU-LSTM-CNN & No & Yes          \\ \hline
		\end{tabular}
	\end{table*}
	
	Each team in the top-5 teams on both public-test and private-test were qualified to submit papers describing the predictive model to the VLSP workshop. There were five submitted papers from five teams including (1) \emph{SunBear}, (2) \emph{ABCD}, (3) \emph{Try hard}, (4) \emph{HH\_UIT}, and (5) \emph{TIN HUYNH}. The predictive models are described in Table~\ref{table2:approaches}. It can be seen that although deep learning works well on the public-test data, conventional feature-based machine learning works better on the private-test data. Furthermore, all the top-3 performing models on the private-test data utilised ensemble learning. This is not a new phenomenon, however, we would like to re-confirm that ensemble learning is applicable for the HSD task in Vietnamese as well.
	
	In particular, the \emph{SunBear} team proposed to utilise logistic regression, a conventional feature-based machine learning model, to handle the task. They used the 35,000 most frequent $n-$grams extracted from the dataset as the input features for training the model. An ensemble learning was then employed to achieve the best macro-average F1 score of 61.97\% on the private-set which is 3\% absolute higher than that produced by the \emph{ABCD} team, the second performing one. They also showed the data pre-processing or normalisation played a very important role in the success of their model as the data from SNSs contains many abbreviations and typos which need to be handled well before training the model. Similarly, the second best performing team (\emph{ABCD}) with macro-averaged F1 of 58.88\% on the private-test data also employed stacking ensemble learning on the outputs of logistic regression models. In their proposed model, many feature types were used including $n-$grams of words, part-of-speech tags and numeric features. 
	
	The remaining three teams employed deep learning to handle the classification problem and achieved good performances on both the public-test and private-test data, especially on the public-test. To the success of the models, all the proposed deep learning models utilised various pre-trained word embeddings which is similar to the findings detailed in the \emph{ETNLP} paper \cite{vu:2019n}. The advantage of deep learning is that there is no need to hand-craft features. While other models did not use word segmentation, the \emph{Try hard} team employed Vietnamese word segmentation \cite{NguyenNVDJ18} on the dataset and achieved the best performance on the public-test and the third performance on the private-test. Moreover, the best performing team on both the public and private test data without ensemble learning is \emph{HH\_UIT}, in which they employed Bi-LSTM with fastText embeddings to handle the task. 
	
	
	\section{Conclusions}
	\label{con}
	The Hate Speech Detection (HSD) shared task in the VLSP
	Campaign 2019 has been a valuable exercise in building
	predictive models to filter out hate speech contents on social networks. It has brought together
	different teams looking at a common goal. We plan to have
	a similar challenge using social network data to better support society in the information age for the next VLSP
	campaign in 2020.


	\section*{Acknowledgment}
	
	The authors would like to thank the InfoRE Technology Company, the team of AiViVN.Com, and the twenty-five annotators for their hard work to support the shared task. Without their support, the task would not have been possible.
	
	
	
	%
	
	
	\bibliographystyle{IEEEtran}
	\bibliography{references}

\end{document}